\documentclass[conference]{IEEEtran}
\IEEEoverridecommandlockouts
\usepackage{cite}
\usepackage{amsmath,amssymb,amsfonts}
\usepackage{hyperref}
\hypersetup{colorlinks,allcolors=black}
\usepackage{lipsum}
\usepackage{graphicx}
\usepackage{textcomp}
\usepackage[table,xcdraw]{xcolor}
\usepackage{tabularx}
\usepackage{xcolor}
\usepackage{algorithm} 
\usepackage{algpseudocode}
\usepackage{amssymb}
\usepackage{float}
\usepackage{caption}
\usepackage{subcaption}
\usepackage[main=english,estonian]{babel}
\usepackage{stfloats}
\usepackage{makecell, booktabs, mathtools}

\def\BibTeX{{\rm B\kern-.05em{\sc i\kern-.025em b}\kern-.08em
    T\kern-.1667em\lower.7ex\hbox{E}\kern-.125emX}}
   
\begin{document}

\title{Energy Pricing in P2P Energy Systems Using Reinforcement Learning
}
\author{Nicolas Avila\textsuperscript{\textsection}, Shahad Hardan\textsuperscript{\textsection}, Elnura Zhalieva\textsuperscript{\textsection}, Moayad Aloqaily, Mohsen Guizani \\ \\

Mohamed Bin Zayed University of Artificial Intelligence (MBZUAI), UAE \\
E-mails: \protect \{nicolas.avila; shahad.hardan; elnura.zhalieva; moayad.aloqaily; mohsen.guizani\}@mbzuai.ac.ae \\

\vspace{-.5cm}

}

\maketitle

\begingroup\renewcommand\thefootnote{\textsection}
\footnotetext{These authors contributed equally to this work}

\begin{abstract}
The increase in renewable energy on the consumer side gives place to new dynamics in the energy grids. Participants in a microgrid can produce energy and trade it with their peers (peer-to-peer) with the permission of the energy provider. In such a scenario, the stochastic nature of the distributed renewable energy generators and the energy consumption increase the complexity of defining the fair prices for buying and selling energy. In this study, we introduce a reinforcement learning framework to help solving this issue by training an agent to set the prices that maximize the profit of all components in the microgrid, aiming to facilitate the implementation of P2P grids in real-life scenarios. The microgrid considers consumers, prosumers, the service provider, and a community battery. Experimental results on the \textit{Pymgrid} dataset shows a successful approach to price optimization for all components in the microgrid. The proposed framework ensures flexibility to account for the interest of these components, as well as the ratio of consumers and prosumers in the microgrid. The results also examine the effect of changing the capacity of the community battery on the profit of the system. The implementation code is available \href{https://github.com/Artifitialleap-MBZUAI/rl-p2p-price-prediction}{here}.
\end{abstract}

\section{Introduction}

Increasing demand for energy due to the population growth and the devastating impact of conventional energy generation sources on global warming spurred a surge of interest in renewable energy sources (RES), mainly solar and wind energy. Worldwide growing investments in RES have encouraged the consumers to locally install rooftop photo-voltaic (PV) systems to lower their electricity bills and make a profit by trading surplus energy within the community. This advancement changed the residential consumers into \textit{prosumers}, who can both cover their demands and provide electricity to other consumers locally.

The use of prosumers' energy alleviated the dependency of the local energy market (LEM) on the utility grid and grew into a peer-to-peer (P2P) energy marketplace. The P2P is a platform where consumers and energy suppliers within or beyond the community microgrid transact energy at the desired price. Eight P2P energy trading pilots were reported around the world in 2020, according to \cite{Irena2020}. These implementations allow the world to recognize the potential benefits of deploying P2P platforms and understand the requirements for their successful realization.

According to the plan of the United Nations and the International Energy Agency \cite{unitednations}, by 2030, countries should aim to reduce 45\% of their carbon emissions, and reach a net zero emissions by 2050. In such a situation, P2P energy markets could help accelerate the execution of the plan by integrating RES into the bulk power grid and leading the energy industry toward a more sustainable direction. However, their implementation is challenging due to the different regulations in the energy sector according to each country. The stochastic nature of RES is another factor that brings complexity.

Nonetheless, the primary challenge behind the realization of P2P platforms lies in agreeing on a bidding strategy that facilitates the trading of energy from prosumers to consumers without affecting the service provider's interests. Developing an optimal dynamic pricing mechanism for P2P energy trading could promote the emergence of more prosumers, and with them, more P2P energy markets.

We aim to tackle the challenges of dynamic pricing in P2P platforms by training a reinforcement learning (RL) agent that learns an optimal pricing policy in real-time. We consider a single microgrid that consists of prosumers and consumers, an energy service provider, the legacy utility grid, and a community battery. The list of main contributions presented in the current work is summarized as follows:

\begin{itemize}
    \item We deal with various stochastic processes in the microgrid, such as demand and generation of customers, to provide a comprehensive problem formulation that depicts the dynamics of a microgrid as accurately as possible.
    \item We adopt the Deep Q-Networks (DQN) method to solve the Markov Decision Process (MDP) problem. The goal is to decide retail and purchase energy prices that minimize the operation cost.
    \item We consider the nature of the microgrid and the individual interest of its members to evaluate the sustainability and flexibility of our framework.
\end{itemize}

The paper follows this structure: Section \ref{rl} introduces and compares the related works on dynamic pricing mechanisms. Section \ref{prob-formulation} provides the MDP formulation for the dynamics of the pricing problem for P2P energy trading platforms. After that, we present the details about using the DQN algorithm for determining optimal retail and purchase prices in section \ref{proposed-solution}. Lastly, in section \ref{eval}, we show the experiments and results. 
\section{Related Work}
\label{rl}
In \cite{8973868}, Paudel and Gooi proposed a pricing strategy for a P2P energy trading platform using the alternating direction method of multiplier (ADMM). They introduced a P2P market operator that offers prices and provides a platform for energy trading from prosumers to consumers. Given the many factors influencing the energy trading price, the authors chose to focus more on the social welfare of the community microgrid. They modeled the personal satisfaction of each household considering consumption and used it to define a function of welfare that depends on the per unit price of energy. The optimization process determines the best energy platform price that maximizes the aggregated welfare of the community microgrid. However, this work did not consider the dynamic nature of energy pricing and provided only a stable solution that does not vary over time.

On the other hand, Kim, Zhang, et al. \cite{7321806} were one of the first to apply RL to allow setting an optimal retail price based on the dynamics of the customer behavior and the change in electricity cost. More specifically, they formulated the dynamic pricing problem as an MDP problem, where a service provider decides the action of choosing a retail energy price at each time step $t$. They defined the cost as the weighted sum of the service provider's cost and the customers' cost at each time step. They solved this MDP problem by adopting a Q-learning algorithm with some proposed improvements. Apart from dynamic pricing, the authors also considered the case where customers can schedule their energy consumption based on the observed energy price to minimize their long-term cost, which turns this problem into a multi-agent learning case. However, the authors did not consider the prosumers' energy generation capability that largely influences the smart-grid dynamics and impacts the retail energy price. Furthermore, the Q-learning algorithm used in this work has high memory space requirements to store the state-action values and takes a long time to converge, making it inefficient to apply with bigger state spaces. Our formulation of the reward function is inspired by this work.

\begin{table*}[bp]
\caption{Summary of Related Work (\checkmark: considered, - : not considered)}
\label{tab:related_works}
\begin{center}
\begin{tabular}{|c|c|c|c|c|c|}
\hline
 & Approach & Price prediction & Real data & \makecell{Prosumers' energy \\ generation capabilities} & Shared battery system   \\
\hline
\cite{8973868} & Optimization (ADMM) & \checkmark & - & \checkmark & -  \\
\hline
\cite{6194235} & Optimization & \checkmark & \checkmark & - & - \\
\hline 
\cite{en13205420}& ML Regression & \checkmark & \checkmark & \checkmark & - \\ 
\hline 
\cite{He_2011}& MDP & \checkmark & - & - & -   \\
\hline
\cite{7321806}& RL (Q-Learning) & \checkmark & \checkmark & - & - \\
\hline
Our work & RL (DQN) & \checkmark & \checkmark & \checkmark & \checkmark  \\
\hline
\end{tabular}
\label{tab:approach}
\end{center}
\end{table*}

Likewise, authors in \cite{He_2011} formulated multi-timescale dispatch and scheduling for a smart-grid model as an MDP problem considering the uncertainty of wind generation and energy demand. Specifically, they proposed the dispatching and pricing in two timescales: real-time and day-ahead scheduling. While the authors made a vast contribution to the integration of wind power into the bulk power grid, they did not consider customers who can generate wind power and actively trade energy with other customers within a smart grid.

Other approaches propose statistical regression models, which identify the set of independent variables required for the complex process of forecasting the electricity price. Authors in \cite{en13205420} argue that there is no fit-for-all set of variables and hence narrowed down their scope by selecting 19 variables based on the characteristics of the UK energy market. They performed a multivariable regression using gradient boosting, random forests, and XGBoost, where the task of each of the models was to make an electricity price forecast 1-12 hours ahead.

Instead of focusing on maximizing social welfare, Joe-Wong, Sen, et al. \cite{6194235} approached the price offerings optimizing problem from the service provider's point of view, maximizing its revenue. By assessing consumers' device-specific scheduling flexibility and modeling their willingness to shift the energy consumption to off-peak periods, the authors formulated an optimization problem to determine cost-minimizing prices for service providers. The authors also argue that real-time pricing is less customer friendly than day-ahead price scheduling since it does not allow the customers to plan their activities in advance and thus creates more uncertainty.

In Table \ref{tab:related_works}, we show the comparison of our proposed framework with the previous studies on dynamic pricing mechanisms for smart grid scenarios.

\section{Problem Formulation}
\label{prob-formulation}
In this work, we define a microgrid composed of a service provider (SP), a set of prosumers $P$, a set of consumers $C$, and a community battery. We consider a temporally dynamic microgrid, where at each time step $t$, the SP adopts a retail energy price $a^t: \mathbb{R}_{+} \mapsto \mathbb{R}_{+}$ and a purchase energy price $p^t: \mathbb{R}_{+} \mapsto \mathbb{R}_{+}$. SP uses $a^t$ to charge both consumers and prosumers depending on their total load demand and uses $p^t$ to calculate how much it has to pay to the prosumers for their energy surplus. In other words, SP regulates both the price to sell energy and the purchase price for which it buys surplus energy from prosumers. Furthermore, SP can also purchase the microgrid's energy requirements from the utility grid (UG) using a fixed cost function. We also consider a shared community battery that facilitates energy trading within the microgrid by storing the surplus energy and partially covering the customers' demands when requested.


We assume that the set of retail pricing functions' and the set of purchase pricing functions' coefficients are both finite, and define them as $\mathcal{A} = \{a_1, a_2, \dots, a_A\}$ and $\mathcal{P} = \{p_1, p_2, \dots, p_P\}$, respectively. We also assume that the demand rate of the customers and the prosumers' renewable (PV) energy generation rate can differ depending on the period of the day. To take into account this temporal dependency, we propose a set of periods $\mathcal{H} = \{0,1,\dots, H-1\}$, where $H$ can be chosen to be $24$. We denote the period of the day at time step $t$ as $h^t$, where $h \in \mathcal{H}$. In the following sections we use the term customer for both consumers and prosumers when they share the same behavior models. In any other cases, we refer to them separately.

\subsection{\textit{Model of the Community Battery System}}

We consider a shared battery within the microgrid that can be charged by the surplus energy coming from the prosumers and discharged to meet some of the load demand within a community. The battery has physical constraints that define the power it can charge or discharge at a specific time step, which we denote by $P_{BC}^t$ and $P_{BD}^t$, respectively. These two parameters change depending on the requested energy to store or retrieve from the battery. The estimator for the available stored energy of a battery is known as a state of charge ($SOC$). The formula of $SOC$ at time step $t$ is inspired from \cite{pymgrid}, and defined as

\begin{equation}\label{soc}
    SOC^{t} = SOC^{t-1} + \frac{P_{BC}^{t} \times \eta - P_{BD}^t/\eta}{\Lambda} ~,
\end{equation}

where $\eta$ is the efficiency of the battery and $\Lambda$ is its nominal capacity. The battery $SOC$ operates within a safe interval, and the battery inverter size restricts the charging and discharging power ($P_{BC}^t$ and $P_{BD}^t$). We define these constraints as

\begin{equation}
    SOC_{min} \leq SOC^t \leq SOC_{max}~,
\end{equation}
\begin{equation}
    0 \leq P_{BC}^t \leq P_{BC, max}, \, 0 \leq P_{BD}^t \leq P_{BD, max}~,
\end{equation}
where $SOC_{min}$ and $SOC_{max}$ are the minimum and maximum values the $SOC$ of the battery can reach, and $P_{BC, max}$ and $P_{BD, max}$ are the upper bounds of the charging and discharging power.

We further assume that prosumers will receive a fixed payment of $b_s$ per energy unit when they store in the battery. Likewise, the customers pay a fixed price of $b_p$ per energy unit to use the energy from the battery. Typically, $b_p$ can sometimes be less than $a^t$, and $b_s$ might be greater than $p^t$ depending on the time step. In other words, if a prosumer is not satisfied with the purchase price $p^t$, they may choose to sell the surplus energy to the community battery instead. Similarly, if a consumer decides to reject the retail price $a^t$, they can purchase energy from the battery if available.

\subsection{\textit{Model of Customer's Response}}

At each time step, each customer has a load demand, which defines the total volume of energy they want to consume for their household appliances at that period. We define the load demand of customer $i$ at time step $t$ as $d_i^t \in \mathcal{D}_i$, where $\mathcal{D}_i$ is the set of customer $i$'s load demand values. Additionally, we assume that each prosumer $i$ can generate energy due to the PV systems at each time step. We define it by $g_i^t \in \mathcal{G}_i$, where $\mathcal{G}_i$ is the set of prosumers' PV generation values. To make the prices influence the demand, we assume that a customer reports their load demand to SP only in two cases: \romannumeral 1) if the retail energy price $a^t$ is more appealing than the purchase price from battery $b_p$, or \romannumeral 2) even if $b_p$ is lower than $a^t$, the battery cannot fully provide the customer's load demand. In this work, we consider that the distribution of the customer's new load demand is known and depends on the period of the day.

\subsubsection{\textit{Consumer's cost}}

We determine consumer $i$'s cost at time step $t$ as
\begin{equation}\label{consumer}
    \phi_i^t(d_{i,sp}^t, d_{i,b}^t) = b_p(d_{i,b}^t) + a^t(d_{i,sp}^t) ~,
\end{equation}
where $d_{i,sp}^t$ and $d_{i,b}^t$ are the consumer $i$'s demand from the service provider and community battery, respectively. As previously mentioned, the values for $d_{i,sp}^t$ and $d_{i,b}^t$ change depending on the advantage of either $a^t$ or $b_p$. 

\subsubsection{\textit{Prosumer's cost}}

There are additional subtleties to consider in the prosumer's response model while formulating the cost function. First, since a prosumer $i$ has the generated energy $g_i^t$, they may either have a surplus of energy $g_i^t - d_i^t$ or an energy shortage $d_i^t - g_i^t$ at every time step $t$ (considered as demand). If a prosumer has surplus, they can sell it to either the SP or the battery. If they further decide to sell the surplus to the battery, the surplus might still be more than the maximum charging power of battery $P_{BC}^t$. In this case, they sell $P_{BC}^t$ amount of surplus energy to the shared battery system and trade the rest of the surplus with the SP. Taking into account these assumptions, we designate the prosumer $i$'s cost function at time step $t$ as the following:

\begin{equation}\label{prosumer}
    \begin{aligned}
        \phi_i^t(d_{i,sp}^t, d_{i,b}^t, \omega_{i,sp}^t, \omega_{i,b}^t) & = b_p(d_{i,b}^t) + a^t (d_{i,sp}^t) \\
        & - b_s (\omega_{i,b}^t) - p^t (\omega_{i,sp}^t) ~,
    \end{aligned}
\end{equation}
where $\omega_{i,b}^t$ and $\omega_{i,sp}^t$ are the surplus energy to be sold to the shared battery and to the SP, respectively.

\subsection{Model of the Service Provider's Cost}
The responsibility of the SP is to operate the energy trading market by satisfying the needs of both prosumers and consumers. To this end, in every time step $t$, the SP has to decide the amount of electric energy it needs to buy from the UG. According to our assumptions, the SP purchases energy from the UG either when the prices announced by the SP are more profitable to the customers than the battery price or when the battery cannot fully provide the customers' load demands. We define the total quantity of energy requested from SP at time step $t$ as $d_{sp}^t$. The cost charged to the SP when it buys energy from the UG is calculated based on the cost function $c^t: \mathbb{R}_{+} \mapsto \mathbb{R}_+$. In this work, the agent does not learn $c^t$. We rather choose $c^t$ to be a linear function of the demand and assume its coefficient to be fixed. Finally, the cost of the SP is a function of the total energy it needs to buy from the UG, the total surplus energy it purchases from the prosumers, and the total energy it sells to the customers. Specifically,
\begin{equation}\label{SP_1}
    \begin{aligned}
        & \psi^t(d_{sp}^t, \omega_{sp}^t) = \\
        & c^t \left( \sum_{i \in C \cup P}d_{i,sp}^t \right) + p^t(\sum_{i \in P}\omega_{i,sp}^t) - a^t(\sum_{i \in C \cup P}d_{i,sp}^t) ~,
    \end{aligned}
\end{equation}
where the last term defines the SP's revenue from selling energy to the customers. Equation (\ref{SP_1}) shows that the SP purchases the exact load demand of customers from the UG and sells it back to the customers. Assuming basic trading principles, we chose the value for the coefficient of function $c^t$ to be less than $a^t$. Thus, we define the cost function $c^t$ as
\begin{equation}\label{cost}
    \begin{split}
        c^t \left(\sum_{i \in C \cup P} d_{i,sp}^t\right) = \sigma \times \sum_{i \in C \cup P} d_{i,sp}^t ~,
    \end{split}
\end{equation}
where $\sigma = 0.15$.  

\section{Proposed solution} \label{proposed-solution}


To find the optimal prices in the proposed microgrid, we opted for using RL because it is able to capture the stochasticity of microgrid by observing the states and learn from the environment feedback to refine its decisions. We characterize our dynamic pricing RL problem for the P2P trading as an MDP problem, with states $s^t=(SOC^{t}, {d}_{sp}^t, h^t)$, actions $(a^t,p^t)$, and reward $r^t(s^t, a^t, p^t)$.
 The operation cost is the weighted sum of all microgrid participants' costs and is defined as

\begin{equation}
    \begin{aligned}
        \rho^t(s^t, a^t, p^t) & = (1 - \alpha - \beta) \psi^t (d_{sp}^t,\omega_{sp}^t) + \alpha \sum_{i \in C} \phi^t_i (d_{i,sp}^t, d_{i,b}^t)\\
        & + \beta \sum_{i \in P} \phi^t_i (d_{i,sp}^t, d_{i,sp}^t, \omega_{i,sp}^t, \omega_{i,b}^t)~,  
    \end{aligned}
\end{equation}
where $ 0 \leq \alpha+\beta \leq 1$ are weighting factors that allow the model to focus on one component more than the other while learning. If $\alpha$ is higher, the model finds that minimizing the consumer's cost is more important. Likewise, a change in parameter $\beta$ determines the importance of minimizing prosumer's cost. Consequently, if both $\alpha$ and $\beta$ are small, the model will put more significance on minimizing the service provider's cost. The reward term depends on the profit of the entire system, which can be defined as the negative operation cost of the whole microgrid as follows
\begin{equation}
    r^t(s^t, a^t, p^t) = -\rho^t(s^t, a^t, p^t) ~,
\end{equation}

Now that the states, actions and rewards are set, the stationary policy that maps states to actions (functions for energy pricing) is defined as $\pi: S \rightarrow A$. The purpose of our dynamic energy pricing problem is to obtain the optimal stationary policy $\pi^{*}$ for each state $s \in S$ that maximizes the expected discounted system reward of the full microgrid as in the following MDP problem $\textbf{P}$:
\begin{equation}
    \textbf{P}: \underset{\pi: S \rightarrow A}{\max} E[\sum_{t=0}^{\infty} (\gamma)^t r^t (s^t, \pi(s^t))]~,
\end{equation}
where $0 \leq \gamma < 1$ represents the discount factor that is used to set a boundary to the accumulated sum of rewards, preventing it to grow indefinitely. It also compares the significance of the present system reward with the future system reward. 

\begin{figure*}[ht]
\centering
\includegraphics[width=0.68\textwidth]{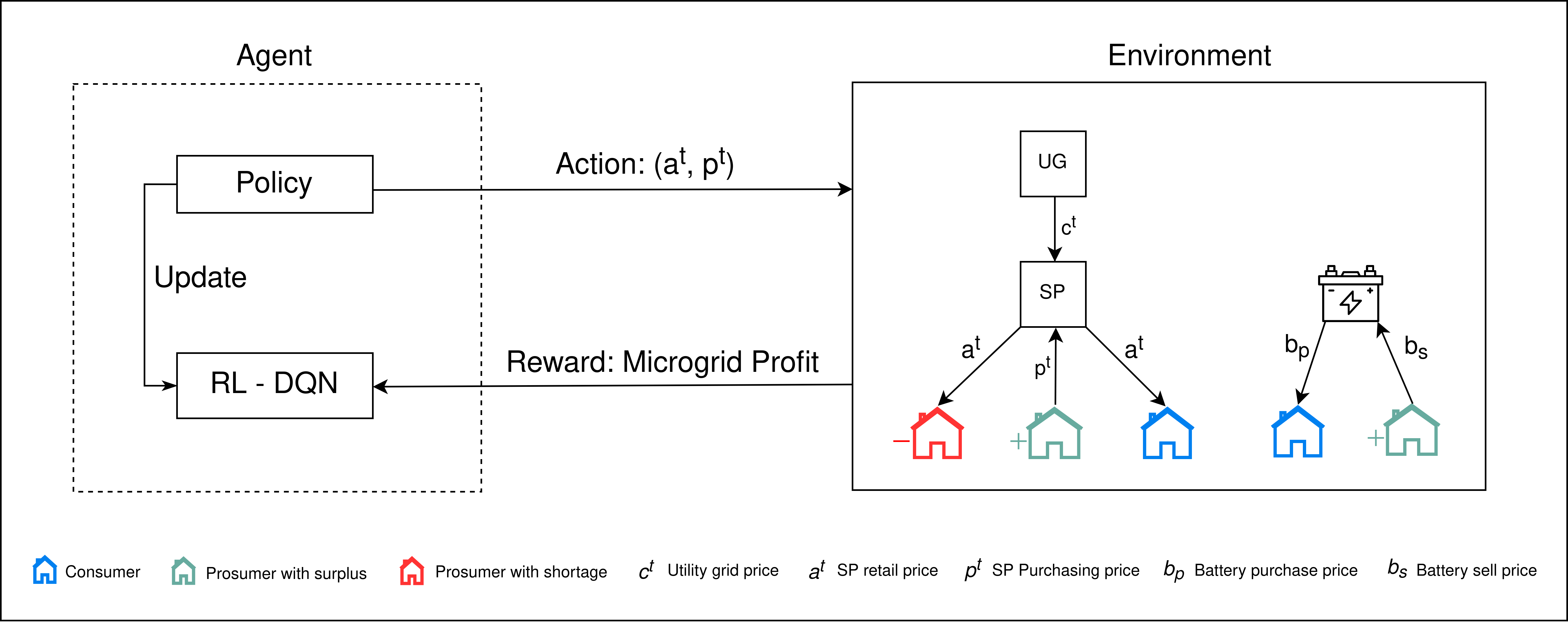}
\caption{System Model: The figure shows the interaction between the components of the environment and the RL agent. The environment contains prosumers who charge the community battery and consumers who discharge it according to their demand. The utility grid provides energy to the SP which then sells it to consumers and prosumers with shortage. Also, the SP buys surplus energy from the prosumers. The RL agent receives the microgrid profit as a reward and returns the retail and purchasing price coefficients.}
\label{sys-graph}
\end{figure*}

We chose a value-based method to solve the presented case: the DQN algorithm introduced by \cite{dqn} and improved in \cite{double_dqn}. It assumes an $\epsilon$-greedy policy that will choose the action that maximizes the expected future reward. A neural network approximates the optimal state-value function $Q^{*}(s,a)= \mathbb{E}[r^t|s^t=s,a^t=a]$. It uses a replay memory to avoid correlation between observations when training $Q$, and an additional target network $\hat{Q}$ that removes the correlation with the target value when calculating the loss. DQN is appropriate for the scenario we propose because we defined a reasonable-sized state space with a discrete action space. Another advantage of DQN is the existence of new contributions that could improve its performance, giving room for improvement for our framework in the future. We present the algorithm functionality in \ref{algo:dqn}, and the interaction between the environment and our RL agent in Fig. \ref{sys-graph}.

\begin{algorithm}
	\caption{Double DQN to learn energy pricing} 
    \label{algo:dqn}
	\begin{algorithmic}[1]
	    \State \textbf{Input:} mini-batch size $K$, steps to learn $E$, experience replay memory size $N$, total time steps $T$, target update interval $U$, learning rate $\zeta$, discount factor $\gamma$
	    \State Initialize $Q, \hat{Q},$ replay memory $D$, $Agent$
	    \For {$step \leftarrow 1, T$}
    	    \Comment{Sample phase}
    	    \State $\epsilon \leftarrow$ setting new epsilon with $\epsilon-$decay
    	    \State Choose an action using $\epsilon-$greedy policy, $a_t \sim \pi_\theta(s_t)$
    	    \State \textit{Agent} takes action $a_t$, observes reward $r_t$, and next state $s_{t+1}$
    	    \State Store the transition ($s_t,a_t,r_t,done_t,s_{t+1}$) in $D$
    	    \If{$step \leq E$}
    	        \Comment Learn phase
    	        \State Sample mini-batch of size $K$ from $D$
    	        \For{$\forall$ ($s_t,a_t,r_t,done_t,s_{t+1}$) $\in D$}
    	            \If{$done_t$}
    	                \State $y_t = r_t$
	                \Else
	                    \State $y_t = r_t + \gamma \max_{a_{t+1} \in A} \hat{Q}(s_{t+1}, a_{t+1})$
                    \EndIf
    	        \EndFor
    	        \State Calculate loss $\mathcal{L}=\frac{1}{K}\sum_{i=0}^{K-1}(Q(s_t,a_t)-y_t)^2)$
    	        \State Update Q using SGD algorithm to minimize $\mathcal{L}$
    	        \If{$step~\%~U \equiv 0$}
    	            \State Copy weights from $Q$ to $\hat{Q}$
    	        \EndIf
	        \EndIf
		\EndFor
	\end{algorithmic} 
\end{algorithm}

\section{Evaluation}
\label{eval}

\subsection{Implementation Setting}



For the discrete action space of DQN, the algorithm chooses values from $\{0.2, 0,4, ...,1 \}$  for $a^t$ and $p^t$, resulting in 25 combinations of coefficients. Following the notation in \ref{algo:dqn}, we set up our experiments using: $K=32$, $\zeta=0.001$, $\gamma=0.99$, $U=1000$, and $T=100K$, with one-year episodes. Regarding the battery, we fixed $b_p=0.3$, $b_s=0.6$, $SOC_{max}=0.9$ and $SOC_{min}=0.1$. Whereas $P_{BC,max}$ and $P_{BD, max}$ are assumed to be 5\% proportional to the battery capacity $\Lambda$. The rest of the hyperparameters changed across the experiments, and the results are shown later in this section.

\subsection{Dataset}

We used the Python library \texttt{Pymgrid} \cite{pymgrid} that is capable of generating and simulating a large number of microgrids. This dataset includes load demand and PV generation data from five cities in the United States with hourly frequency and collected over one year, resulting in 8760 data points. We are interested in the load and PV generation data for initializing our state space. Thus, we only extracted these columns from the \texttt{Pymgrid} dataset. \\
As a pre-processing step, we scaled the values down to the level of participants. For numerical experiments, we consider a microgrid with 10 participants, with an equal number of prosumers and consumers. However, the number of participants is not fixed per se, and can be controlled along with the ratio of consumers to prosumers.

\begin{figure}[ht]
    \centering
    \includegraphics[width=0.35\textwidth]{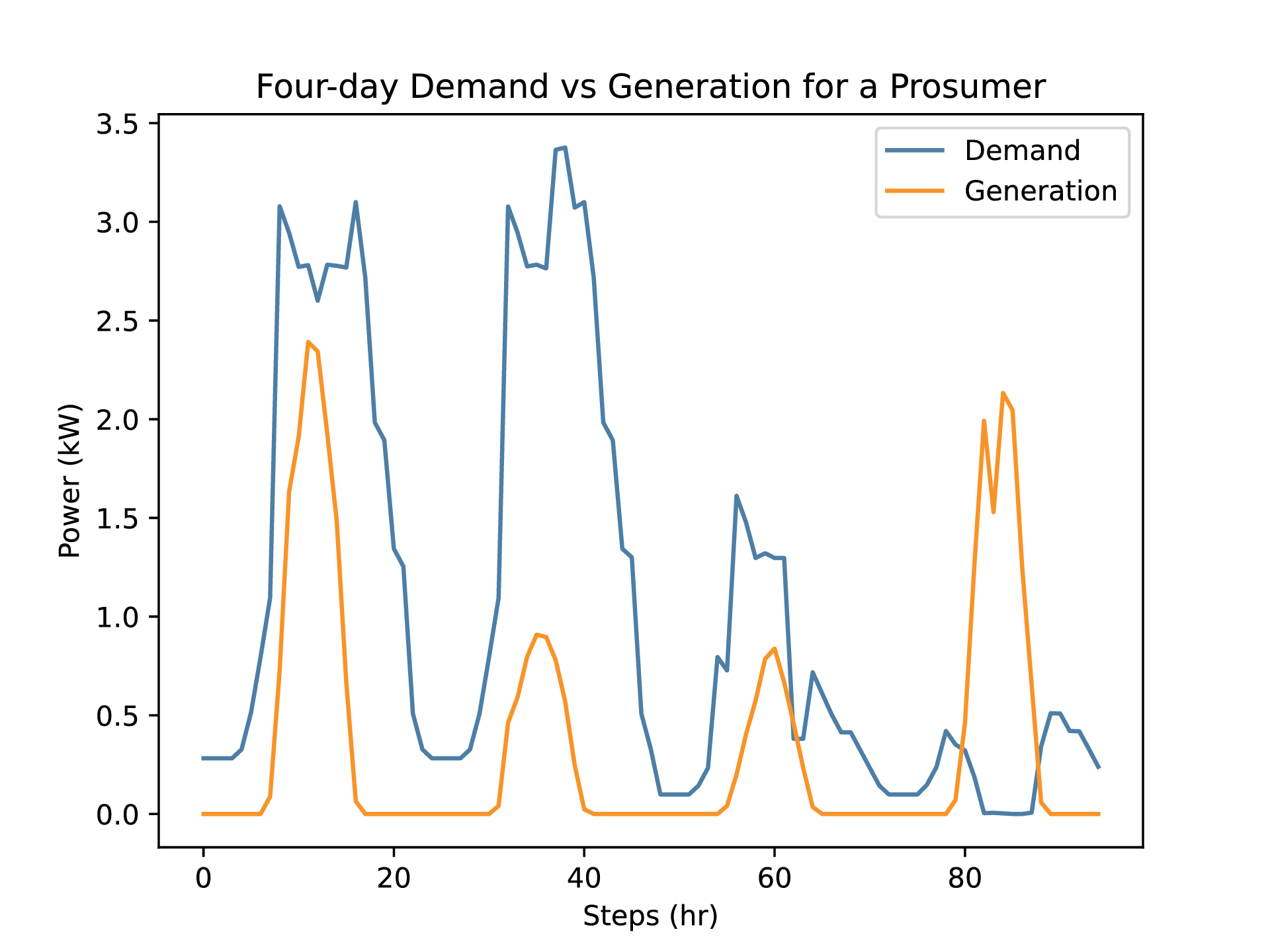}
    \caption{Demand vs. generation for a prosumer in a period of four days.}
    \label{fig:prosumer_demand_vs_gen}
\end{figure}

\subsection{Results}


We analyzed our proposed solution by studying the effect of changing the model's emphasis on the components of the microgrid, the battery capacity, and the consumer/prosumer ratio. To demonstrate the model's effectiveness, we ran an experiment with an equal number of consumers and prosumers and a battery capacity of 30 kWh. To do so, we set $\alpha=0.3$ and $\beta=0.3$, giving $0.4$ to the SP term. Fig. \ref{fig:dqn_learning} shows the decreasing loss and the reward behavior. The oscillations in the reward in Fig. \ref{fig:dqn_reward} are caused by the dynamic nature of the dataset. As the demand and generation given as an input at each time step change hourly, it is expected that the reward will change accordingly, rather than having a monotonic behavior.

Moreover, since the reward function is the negative weighted sum of the components' profit, it provides the ability to prioritize one over the other in the system. We examined the changes in each component's profit upon varying the values of the consumers' $\alpha$, the prosumers' $\beta$, and the SP's $(1-\alpha - \beta)$. Table \ref{change-alpha-beta} presents the results for a microgrid with an equal number of consumers and prosumers and a battery capacity of 30 kWh. It shows that when the focus is higher on the consumers, their profit increases, and the prosumer's profit decreases. The SP profit is significantly affected by the changes in $\alpha$ due to the trade-off caused by changing the hyperparameters simultaneously, even when the SP coefficient is relatively low. The analysis proved that our model is flexible to account for the members' profit in a real-life implementation.

\begin{figure}
     \centering
     \begin{subfigure}[b]{0.24\textwidth}
         \centering
         \includegraphics[width=\textwidth]{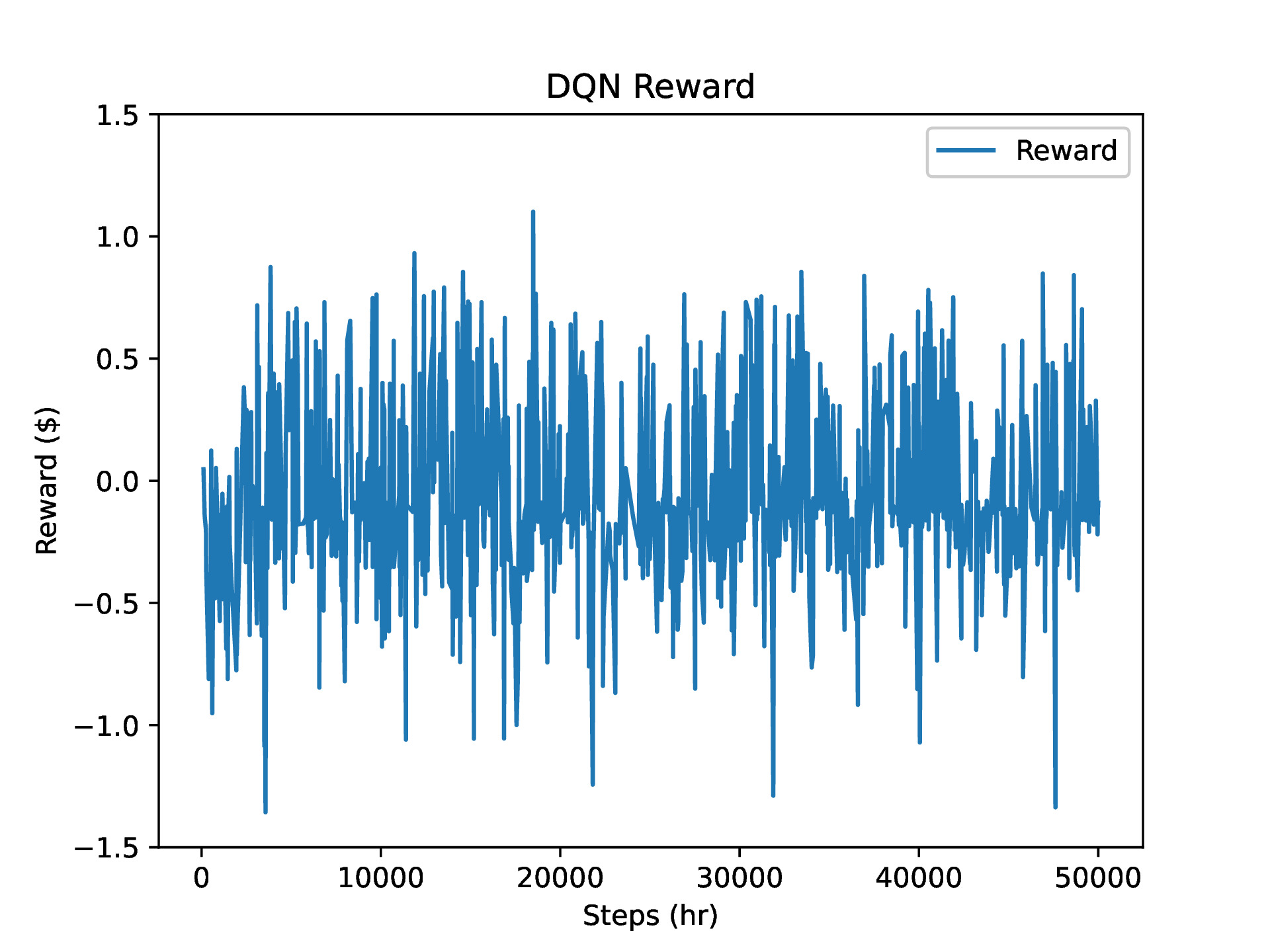}
         \caption{$Reward$}
         \label{fig:dqn_reward}
     \end{subfigure}
     \begin{subfigure}[b]{0.24\textwidth}
         \centering
         \includegraphics[width=\textwidth]{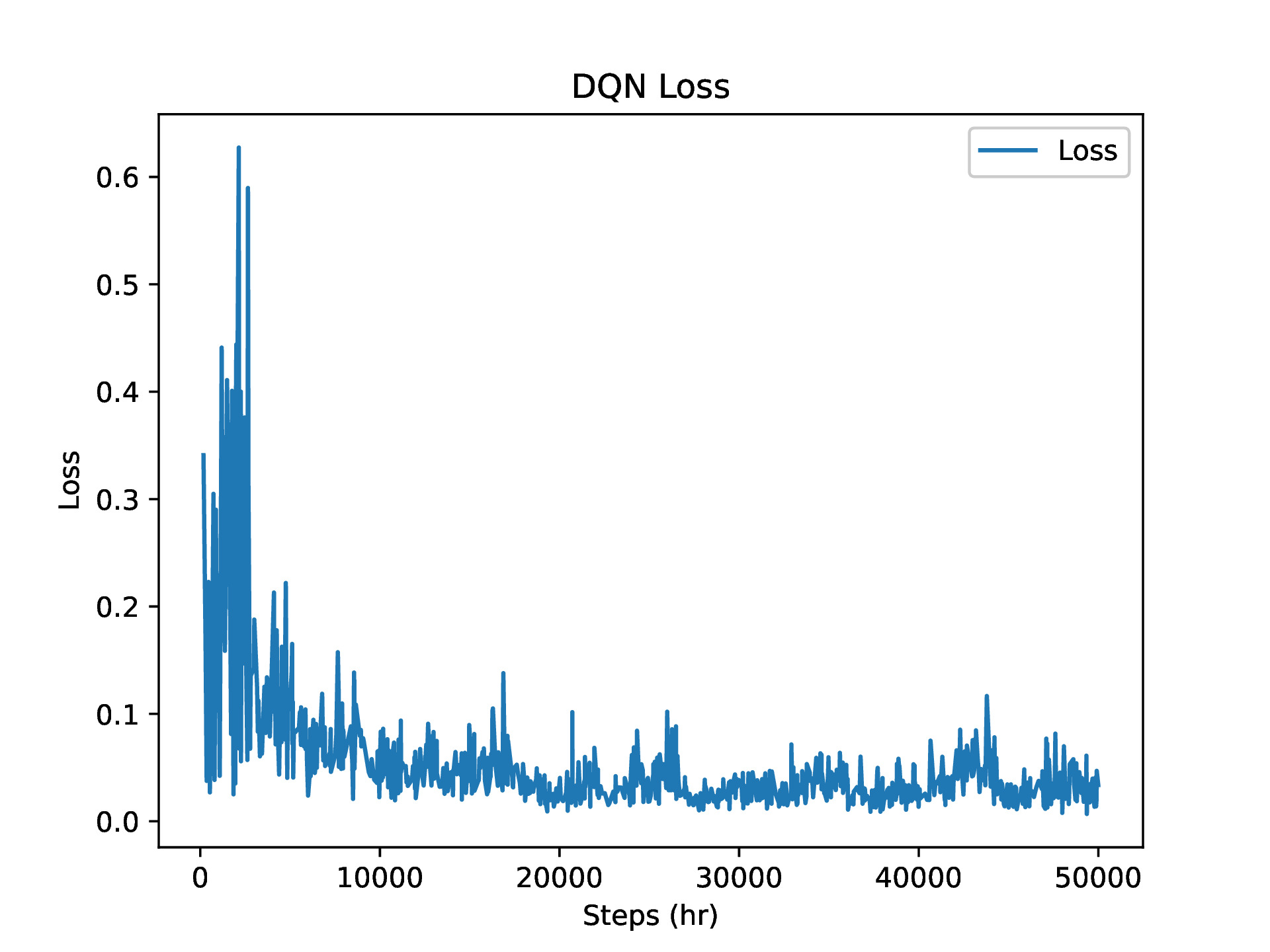}
         \caption{$Loss$}
         \label{fig:dqn_loss}
     \end{subfigure}
        \caption{DQN learning process.}
        \label{fig:dqn_learning}
\end{figure}

We experimented with the battery capacity size and maximum charging/discharging rates to see how they influence the total reward calculated over the last year of training. Fig. \ref{fig:batt_reward} shows the results of these evaluations. The setting defined an equal weighting factor for prosumers and consumers' cost and the highest focus on the SP. As the battery capacity increases, the total reward decreases. The reason is that a higher capacity of the community battery promotes its use among the community. Thus, customers will depend less on the service provider, which has the highest focus of the agent, resulting in a significant decrease in the system's profit. 

\begin{figure}[ht]
    \centering
    \includegraphics[width=0.35\textwidth]{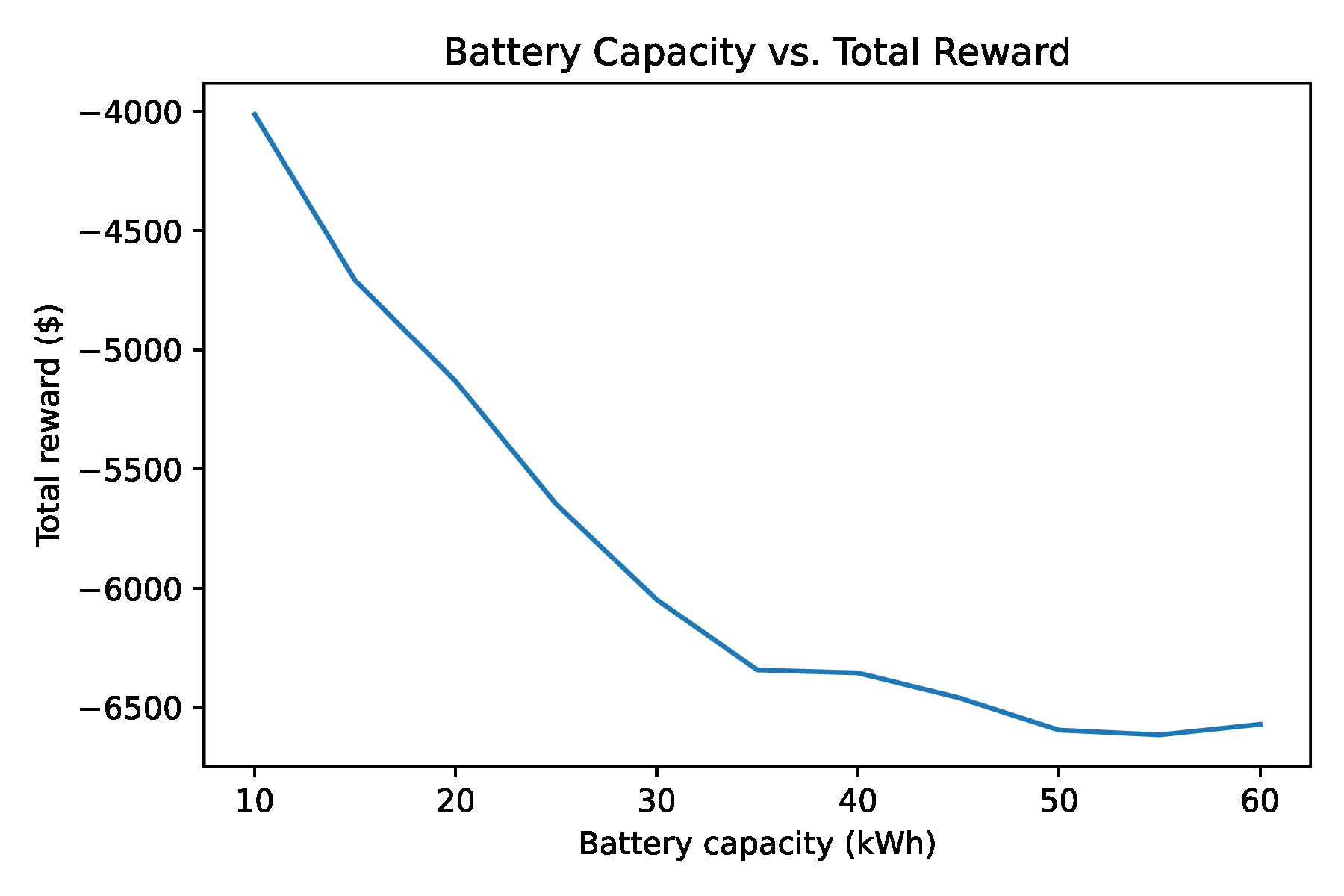}
    \caption{Relationship between the battery capacity and total reward over the last year.}
    \label{fig:batt_reward}
\end{figure}

Lastly, our model shows that as the consumer rate increases, the average reward of the system increases, as shown in Fig. \ref{fig:consumer_rate}. The result is expected because increasing the number of consumers increases the demand from the SP, leading to increasing its profit. Furthermore, prosumers will be less able to effectively charge the battery and provide more affordable energy to the consumers.

\begin{figure}[ht]
    \centering
    \includegraphics[width=0.35\textwidth]{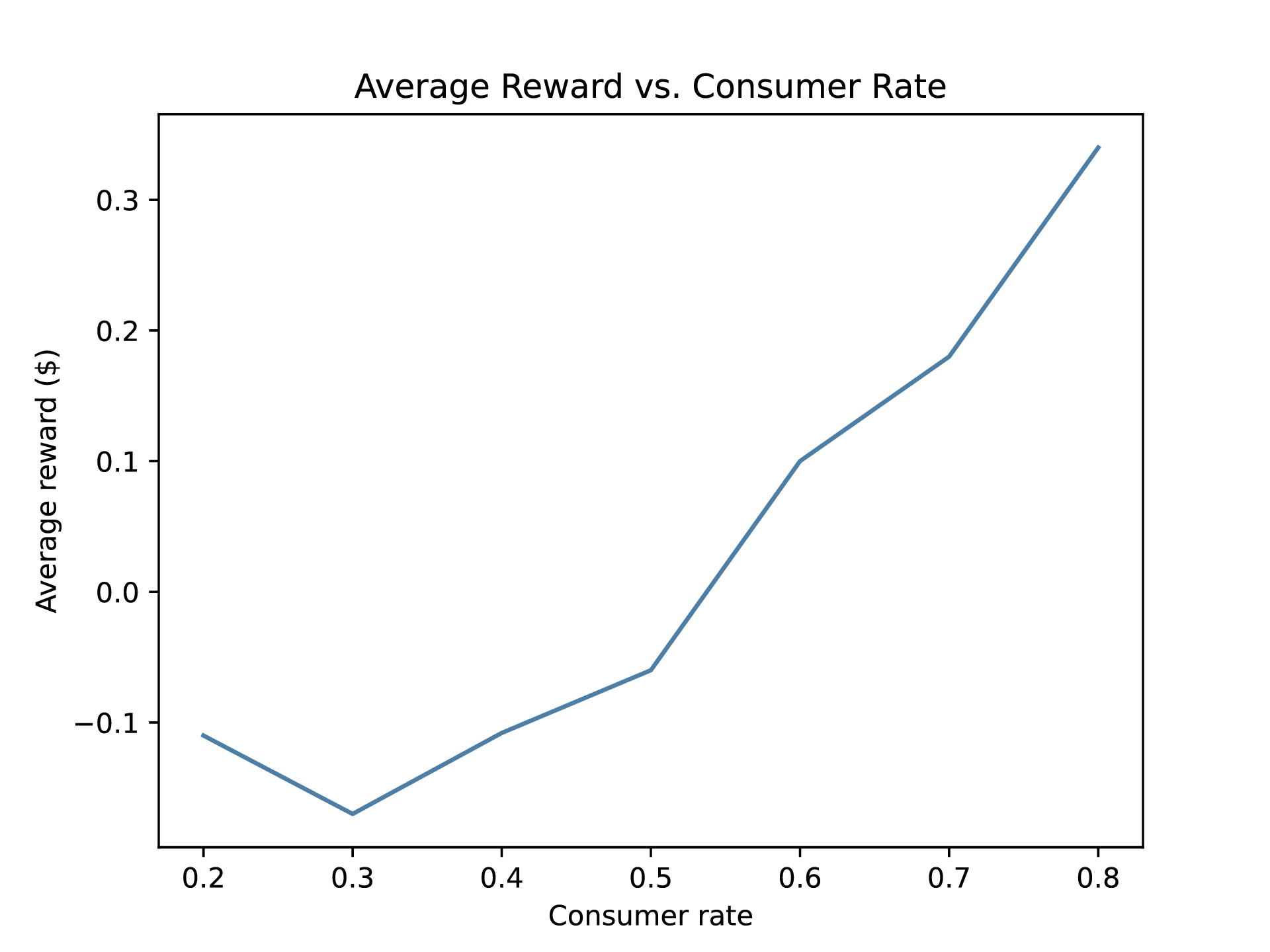}
    \caption{Relationship between the consumer rate and the average reward of the system.}
    \label{fig:consumer_rate}
\end{figure}

\section{Conclusion}\label{conclusion}

The continuous efforts to promote more sustainable energy consumption can be supported by involving the citizens in the energy market. Our work shows that RL can optimize the interest of consumers, prosumers, and the service provider. The proposed framework proves its success in considering the order of priority between the components. We found that including a community battery increases the independence of customers with the appropriate battery capacity. The proposed setting could be improved by better modelling the dynamic demand of the customers and the PV generation. Furthermore, we could implement concepts such as demand response for generating customers data to encourage actions from the consumers' side according to the price. The DQN algorithm could be replaced by other RL approaches, such as policy-based methodologies.

\begin{table}[htbp]
\caption{Changes of average consumer, prosumer, and SP cost as a result of changing weights of these components in the cost function.}
\begin{center}
\begin{tabular}{|c|c|c|c|c|c|}
\hline
\thead{$\alpha$}& \thead{$\beta$}& \thead{$1-\alpha-\beta$}& \thead{Avg. C.\\profit}& \thead{Avg. P.\\profit} & \thead{Avg. SP\\profit}  \\
\hline
0.2& 0.6& 0.2& -8.83& 1.24& 5.02 \\
\hline
0.2& 0.2& 0.6& -9.05& 0.21& 6.00\\
\hline
0.6& 0.2& 0.2& -2.78& -0.08& 0.64  \\
\hline
0.3& 0.3& 0.4& -9.40& 0.22& 6.72 \\
\hline
0.3& 0.5& 0.2& -4.90& 1.00 & 1.40 \\
\hline 
0.5& 0.3& 0.2& -2.98& 0.53& 0.12\\ 
\hline
0.5& 0.2& 0.3& -2.87& -0.32& 0.98 \\
\hline
0.4& 0.4& 0.2& -3.12& 0.20& 0.67 \\ 
\hline
0.1& 0.7& 0.2& -9.07& 1.07& 5.45 \\ 
\hline 
0.7& 0.1& 0.2& -2.71& -0.138& 0.62 \\
\hline
\end{tabular}
\label{change-alpha-beta}
\end{center}
\end{table}

\section{Acknowledgments}
We thank Dr. Martin Taká\v{c}, an associate professor at MBZUAI, for his valuable input that improved the formulation of our proposed framework.


{
\small
\bibliographystyle{unsrt}
\bibliography{ref.bib}
}

\vspace{12pt}

\end{document}